\documentclass[10pt,twocolumn,letterpaper]{article}

\usepackage{cvpr}
\usepackage{times}
\usepackage{epsfig}
\usepackage{graphicx}
\usepackage{amsmath}
\usepackage{amssymb}

\usepackage{adjustbox}
\newcommand{\widthscalefive}{0.120}
\newcommand{\widthscalefour}{0.160}
\usepackage{animate}

\usepackage{colortbl}
\usepackage{booktabs}
\usepackage[table,x11names,dvipsnames,xcdraw]{xcolor}

\usepackage[pagebackref=true,breaklinks=true,letterpaper=true,colorlinks,bookmarks=false]{hyperref}

\cvprfinalcopy % *** Uncomment this line for the final submission

\def\cvprPaperID{5932} % *** Enter the CVPR Paper ID here

\ifcvprfinal\fi
\begin{document}

%%%%%%%%% TITLE
\title{Bringing Alive Blurred Moments}

\author{Kuldeep Purohit$^{1}$ \qquad Anshul Shah$^{2}\thanks{Work done while at Indian Institute of Technology Madras, India.}$ \qquad A. N. Rajagopalan$^{1}$ \\
$^1$ Indian Institute of Technology Madras, India \hspace{1cm}
$^2$ University of Maryland, College Park\\
%{\tt\small \{chi.zhang,f.gao,baoxiongjia,yixin.zhu\}@ucla.edu, sczhu@stat.ucla.edu}
{\tt\small kuldeeppurohit3@gmail.com, anshulb@cs.umd.edu, raju@ee.iitm.ac.in}
} 

%%\author{Kuldeep Purohit, Anshul Shah, and A N Rajagopalan\\
%%Indian Institute of Technology Madras, India\\
%%%Institution1 address\\
%%{\tt\small  \{ee14s007,ee13b068,raju\}@ee.iitm.ac.in}}

%\and

%\author{Kuldeep Purohit, Anshul Shah, A N Rajagopalan}
%\institute{Indian Institute of Technology Madras, India\\
%	\email{ \{ee14s007,ee13b068,raju\}@ee.iitm.ac.in}
%}

\maketitle

\begin{abstract}
We present a solution for the goal of extracting a video from a single motion blurred image to sequentially reconstruct the clear views of a scene as beheld by the camera during the time of exposure. We first learn motion representation from sharp videos in an unsupervised manner through training of a convolutional recurrent video autoencoder network that performs a surrogate task of video reconstruction. Once trained, it is employed for guided training of a motion encoder for blurred images. This network extracts embedded motion information from the blurred image to generate a sharp video in conjunction with the trained recurrent video decoder. As an intermediate step, we also design an efficient architecture that enables real-time single image deblurring and outperforms competing methods across all factors: accuracy, speed, and compactness. Experiments on real scenes and standard datasets demonstrate the superiority of our framework over the state-of-the-art and its ability to generate a plausible sequence of temporally consistent sharp frames. 
\end{abstract}

\section{Introduction}

When shown a motion blurred image, humans can mentally reconstruct (sometimes ambiguously perhaps) a temporally coherent account of the scene that represents what transpired during exposure time. However, in computer vision, natural video modeling and extraction has remained a challenging problem due to the complexity and ambiguity inherent in video data. With the success of deep neural networks in solving complex vision tasks, end-to-end deep networks have emerged as incredibly powerful tools. 

Recent works on future frame prediction reveal that direct intensity estimation leads to blurred predictions. Instead, if a frame is reconstructed based on the original image and corresponding transformations, both scene dynamics and invariant appearance can be preserved well. Based on this premise, \cite{flynn2016deepstereo,zhou2016view} and \cite{liu2017video} model the task as a flow of image pixels. The methods \cite{vondrick2016generating,xue2016visual} generate a video from a single sharp image, but have a severe limitation in that they work only on the specific scene for which they are trained. All of these approaches work only on sharp images and videos. However, motion during exposure is known to cause severe degradation in the captured image quality due to the blur it induces. This is usually the case in low-light situations where the exposure time of each frame is high and in scenes where significant motion happens within the exposure time. In \cite{vasiljevic2016examining}, it has been shown that standard network models used for vision tasks and trained only on high-quality images suffer a significant degradation in performance when applied to images degraded by blur.

Motion deblurring is a challenging problem in computer vision due to its ill-posed nature. Recent years have witnessed significant advances in deblurring \cite{vasu2017local,pan2016blind,pan2014deblurring}. Several methods \cite{xu2010two,pan2016robust,fergus2006removing,shan2008high,cho2009fast,joshi2008psf,krishnan2009fast,krishnan2011blind,xu2010two} have been proposed to address this problem using hand-designed priors as well as Convolutional Neural Networks (CNN)  \cite{chakrabarti2016neural,schuler2013machine,schuler2016learning}  for recovering the latent image. A few methods \cite{sun2015learning,gong2017motion} have been proposed to remove heterogeneous blur but they are limited in their capability to handle general dynamic scenes. Most of these methods strongly rely on the accuracy of the assumed image degradation model and include intensive, sometimes heuristic, parameter-tuning and expensive computations, factors which severely restrict their accuracy and applicability in real-world scenarios. The recent works of \cite{nah2017deep,nimisha2017blur,kupyn2017deblurgan,tao2018scale} overcome these limitations to some extent by learning to directly generate the latent sharp image, without the need for blur kernel estimation.

We wish to highlight here that until recently, all existing methods were limited to the task of generating only $\lq$a' deblurred image. In this paper, we address the task of reviving and reliving all the sharp views of a scene as seen by the camera during its flight within the exposure time. Recovering sharp content and motion from motion blurred images can be valuable for uncovering the underlying dynamics of a scene (e.g., in sports, traffic surveillance monitoring, entertainment etc.). The problem of extracting a video from a single blurred observation is challenging due to the fact that a blurred image can only reveal aggregate information about the scene during exposure. The task requires recovery of sharp frames which are temporally and scene-wise consistent in the sense that they emulate recording coming from a high frame-rate camera. State-of-the-art deblurring methods such as \cite{vasu2017local} \cite{pan2016blind} estimate at best a group of poses which constitute the camera motion, but with total disregard to their ordering. For example, one would get the same blurred image even if the temporal order is reversed (temporal ambiguity). As a post-processing step, synthesizing a sequence from this group of poses is a non-trivial task. Although the camera motion can be partially detected through gyroscope sensors attached to modern cameras, the obtained data is too sparse to completely describe trajectories within the time interval of a single lens exposure. More importantly, sensor information is seldom available for most internet images. Further, these methods can only handle blur induced by a camera imaging a static planar scene which is not representative of a typical real-world scenario and hence not very interesting.

We present a two-stage deep convolutional architecture to carve out a video from a motion blurred image that is applicable to non-uniform motion caused by individual or combined effects of camera motion, object motion and arbitrary depth variations in the scene. We avoid overly simplified models to represent motion and hence refrain from creating synthetic datasets for supervised training. The first stage consists of training a video auto-encoder wherein the encoder accepts a sequence of video frames to extract a latent motion representation while the decoder estimates the same video by applying estimated motion trajectories to a single sharp frame in a recurrent fashion. We use this trained video decoder to guide the training of a CNN (which we refer to as Blurred Image Encoder (BIE)) to extract the same motion information from a blurred image as the video encoder would from the image sequence corresponding to that blurred image. For testing, we propose an efficient deblurring network to first estimate a sharp frame from the given blurred image. The BIE is responsible for extracting motion features from the blurred image. The video decoder uses the outputs of the BIE and the deblurred sharp frame to generate the video underlying the motion blurred image.

As the only other work of this kind, \cite{jin2018learning} very recently proposed a method to estimate a video from a single blurred image by training multiple neural networks to estimate the underlying frames. In contrast, our architecture utilizes a single recurrent neural network to generate the entire sequence. Our recurrent design implicitly addresses temporal ambiguity to a large extent, since generation of any frame in the sequence is naturally preconditioned on all the previous frames. The approach of \cite{jin2018learning} is limited to small motion, owing to its architecture and training procedure. We estimate pixel level motion instead of intensities which proves to be an advantage for the task at hand, especially in cases with large blur (which is an issue with \cite{jin2018learning}). Our deblurirng architecture not only outperforms all existing deblurring methods but is also smaller and significantly faster. In fact, separating the processes of content and motion estimation allows our architecture to be used with any off-the-shelf deblurring approach.

Our work advances the state-of-the-art in many ways. The main contributions are:

\begin{itemize}
\item A novel solution for extracting a sharp video from a single motion blurred image. In comparison to the state-of-the-art~\cite{jin2018learning}, our network is faster, more accurate (especially for large blur) and contains fewer parameters.

\item A two-stage training strategy with a recurrent architecture for learning to extract an ordered spatio-temporal motion representation from a blurred image in an unsupervised manner. Unlike \cite{jin2018learning}, our network is independent of the number of frames in the sequence.

\item An efficient architecture to perform real-time single image deblurring that also delivers superior performance over the state-of-the-art in deburring~\cite{tao2018scale} across all factors: accuracy, speed ($20$ times faster) and compactness.

\item Qualitative and quantitative analysis using benchmark datasets to demonstrate the superiority of our framework over competing methods in deblurring as well as video generation from a single blurred image. 

\end{itemize}

\section{The Proposed Architecture}
\label{sec:proposed architecture}

\begin{figure*}
\begin{center}
   \includegraphics[width=0.8\linewidth]{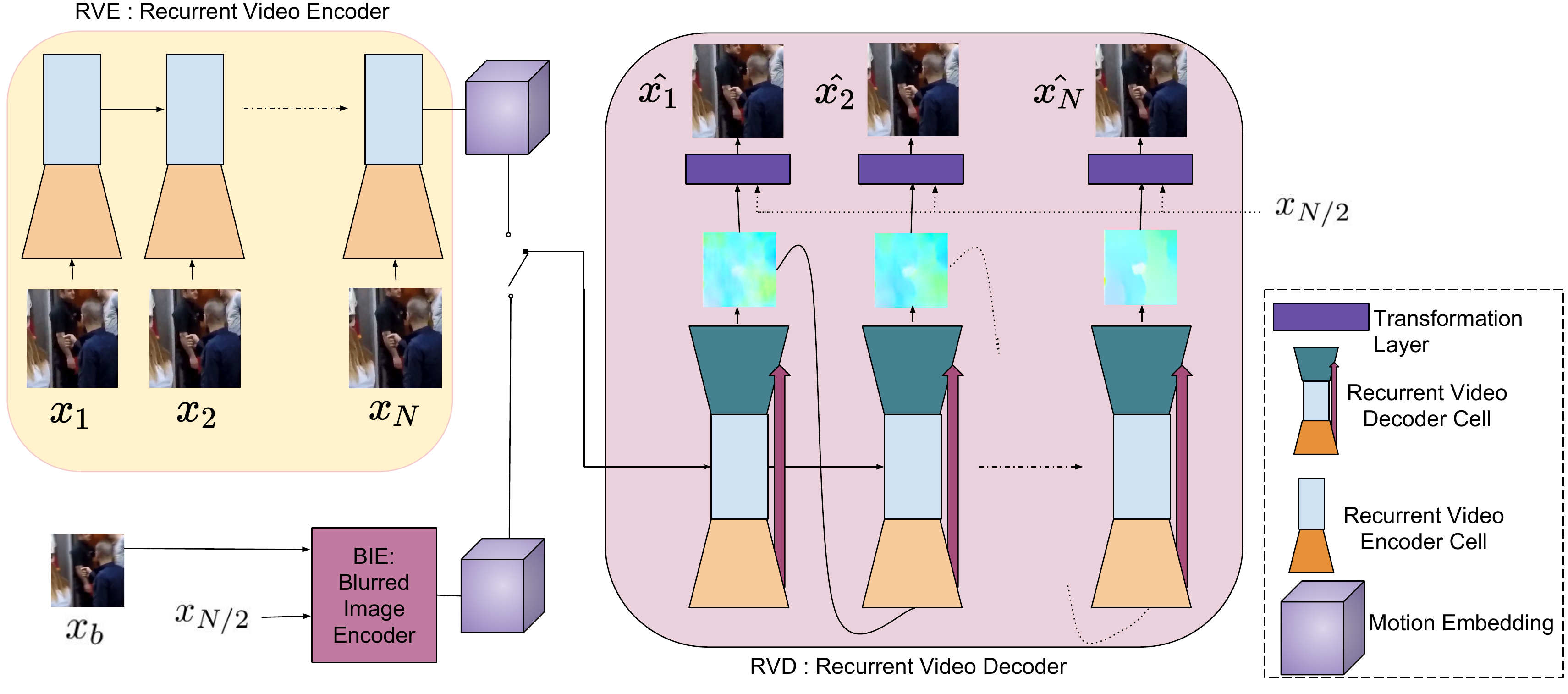}
\end{center}
   \caption{An overview of our video generation architecture during training. The first step involves training the RVE-RVD for the task of video reconstruction. This is followed by guided training of BIE through the trained RVD.}
\label{fig:architecture}
\end{figure*}

Convolutional neural networks (CNNs) have been successfully applied for various vision tasks on images but translating these capabilities to video is non-trivial due to their inefficiency in exploiting temporal redundancies present in videos. Recent developments in recurrent neural networks provide powerful tools for sequence modeling as demonstrated in speech recognition \cite{graves2013speech} and caption generation for images \cite{vinyals2015show}. Long  short term memory networks (LSTMs) can be used to generate outputs that are highly correlated along the temporal dimension and hence form an integral part of our video generation framework. Though Conv3Ds have been used for video classification approaches, we found that for our application, recurrent networks were more efficient. Considering that we are working with images, the spatial information across the image is equally important. Hence we use Convolutional LSTM units \cite{xingjian2015convolutional} as our building blocks, which are capable of capturing both spatial and temporal dependencies.

The task of generating an image sequence requires the network to understand and efficiently encode static as well as dynamic information for a certain period of time. Although such an encoding is not clearly defined and hence unavailable in labeled datasets, we overcome this challenge by unsupervised learning of motion representation. We propose to use video reconstruction as a surrogate task for training our BIE. Our hypothesis is that a successful solution to the video reconstruction task will allow a video autoencoder to learn a strong and meaningful motion representation which will enable it to impart spatio-temporal coherence to the generated moving scene content.

In our proposed video autoencoder, the encoder utilizes all the video frames to extract a latent representation, which is then fed to decoder which estimates the frame sequence in a recurrent fashion. The Recurrent Video Encoder (RVE) reads $N$ sharp frames ${x}_{1..N}$, one at each time-step. It returns a tensor at the last time-step, which is utilized as the motion representation of the image sequence. This tensor is used to initialize the first hidden state of another ConvLSTM based network called Recurrent Video Decoder (RVD) whose task is to recurrently estimate $N$ optical flows. Since the RVE-RVD pair is trained using reconstruction loss between the estimated frames $\hat{x}_{1..N}$ and ground-truth frames ${x}_{1..N}$, the RVD must return the predicted video. To enable this, the (known) central frame of the video is acted upon by the flows predicted by the RVD. Specifically, the estimated flows are individually fed to a differentiable transformation layer to transform the central frame $x_{\lfloor\frac{N}{2}\rfloor}$ to obtain the frames $\hat{x}_{1..N}$. Once trained, we have an RVD which can estimate sequential motion flows, given a particular motion representation.

In addition, we introduce another network called Blurred Image Encoder (BIE) whose task is to accept blurred image $x_B$ corresponding to the spatio-temporal average of the input frames ${x}_{1..N}$ and return a motion encoding, which too can be used to generate a sharp video.  
To achieve this task, we employ the already trained RVD to guide the training of BIE so as to extract the same motion information from the blurred image as the RVE would from that image sequence. In other words, the weights are to be learnt such that $BIE(x_B) \approx RVE(x_{1..N})$. We refrain from using the encoding returned by RVE for training due to lack of ground truth for the encoded representation. Instead, the BIE is trained such that the predicted video at the output of RVD for the given $x_B$ matches as closely as possible to the ground truth frames ${x}_{1..N}$. This ensures that the BIE learns to capture ordered motion information for the RVD to return a realistic video. Directly training the BIE-RVD pair poses a challenge since it requires learning to perform two tasks jointly: ``video generation from motion representation'' and ``ambiguity-invariant motion extraction from a blurred image''. Such training delivers below-par performance (see supplementary material).% Directly training the BIE-RVD pair is cumbersome and may lead to the RVD learning to extract motion whereas this task is to be performed by RVE/BIE.

The overall architecture of the proposed methodology is given in Fig. \ref{fig:architecture}. It is fully convolutional, end-to-end differentiable and can be trained using unlabeled high frame-rate videos, without the need for optical flow supervision, which is challenging to produce at large scale. During \emph{testing}, the central sharp frame is not available and is estimated using an independently trained deblurring module (DM). We now describe the design aspects of the different modules.

\subsection{Recurrent Video Encoder (RVE)}
\label{sec:rve}

\begin{figure}
\begin{center}
 \begin{tabular}{cc}
   \includegraphics[width=0.24\textwidth]{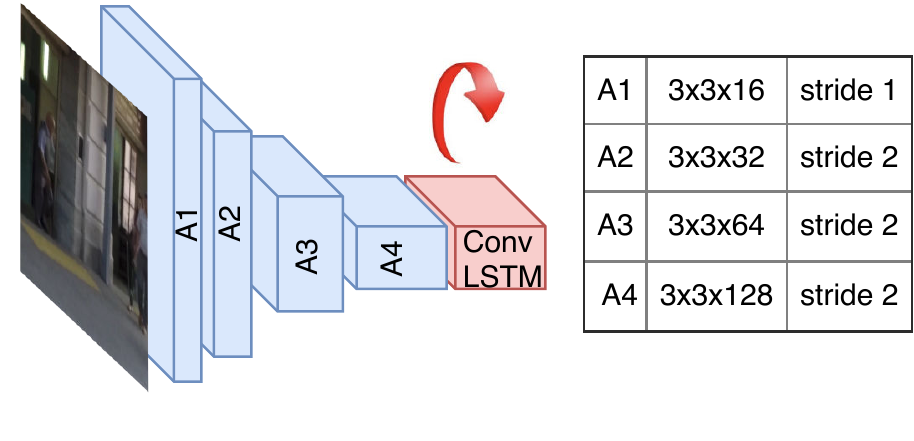} &
      \includegraphics[width=0.20\textwidth]{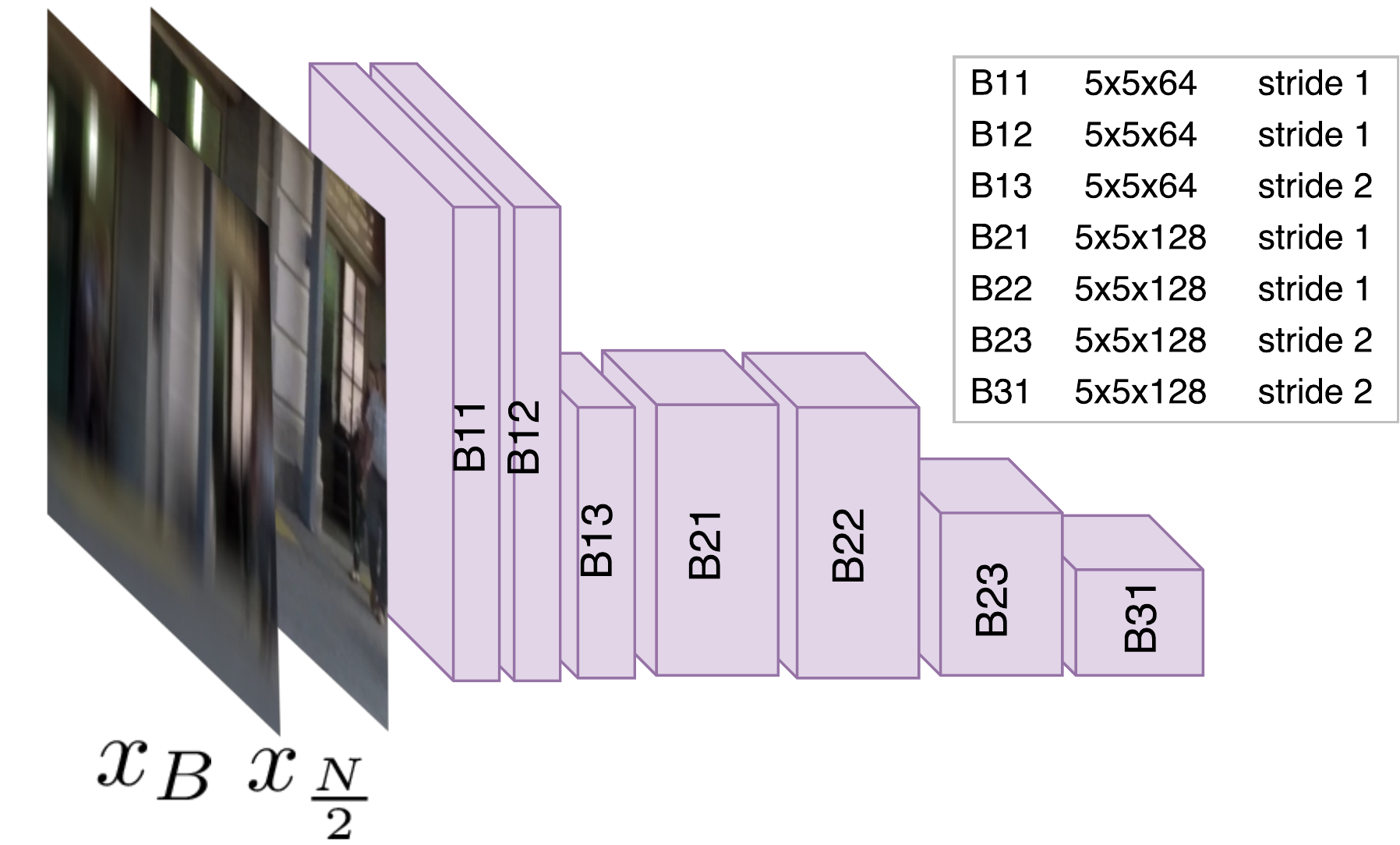}\\

            (a) RVE architecture. & (b)  BIE architecture.
       \end{tabular}
\end{center}
   \caption{Architectures of BIE and RVE. The RVE is trained to extract a motion representation from a sequence of frames while the BIE is trained to extract a motion representation from a blurred image and a sharp image.}
\label{fig:BIEnRVE}
\end{figure}

At each time-step, a frame is fed to a convolutional encoder, which generates a feature-map to be fed as input to the ConvLSTM cell. Interpreting ConvLSTM's hidden-states as a representation of motion, the kernel-size of a ConvLSTM is correlated with the speed of the motion which it can capture. Since we need to extract motion taking place within a single exposure at fine resolution, we choose a kernel-size of $3\times3$. As can be seen in Fig. \ref{fig:BIEnRVE}(a), the encoder block is made of $4$ convolutional blocks with $3\times3$ filters. The first block is a conv layer with stride of $1$ and the rest contain a conv layer with stride of $2$, followed by a Resblock. The number of feature maps in the outputs of these blocks are $16$, $32$, $64$ and $128$, respectively. A ConvLSTM cell operates on the features returned by the last block and augments it with memory from previous time-steps.

Overall, each module can be represented as $h^{enc}_n = enc(h^{enc}_{n-1},x_{n})$,
where $h^{enc}_{n}$ is encoder ConvLSTM state at time step $n$ and $x_{n}$ is the $n^{th}$ sharp frame of the video.

\subsection{Recurrent Video Decoder (RVD)}
\label{sec:rvd}

\begin{figure}
\begin{center}
   \includegraphics[width=0.45\textwidth]{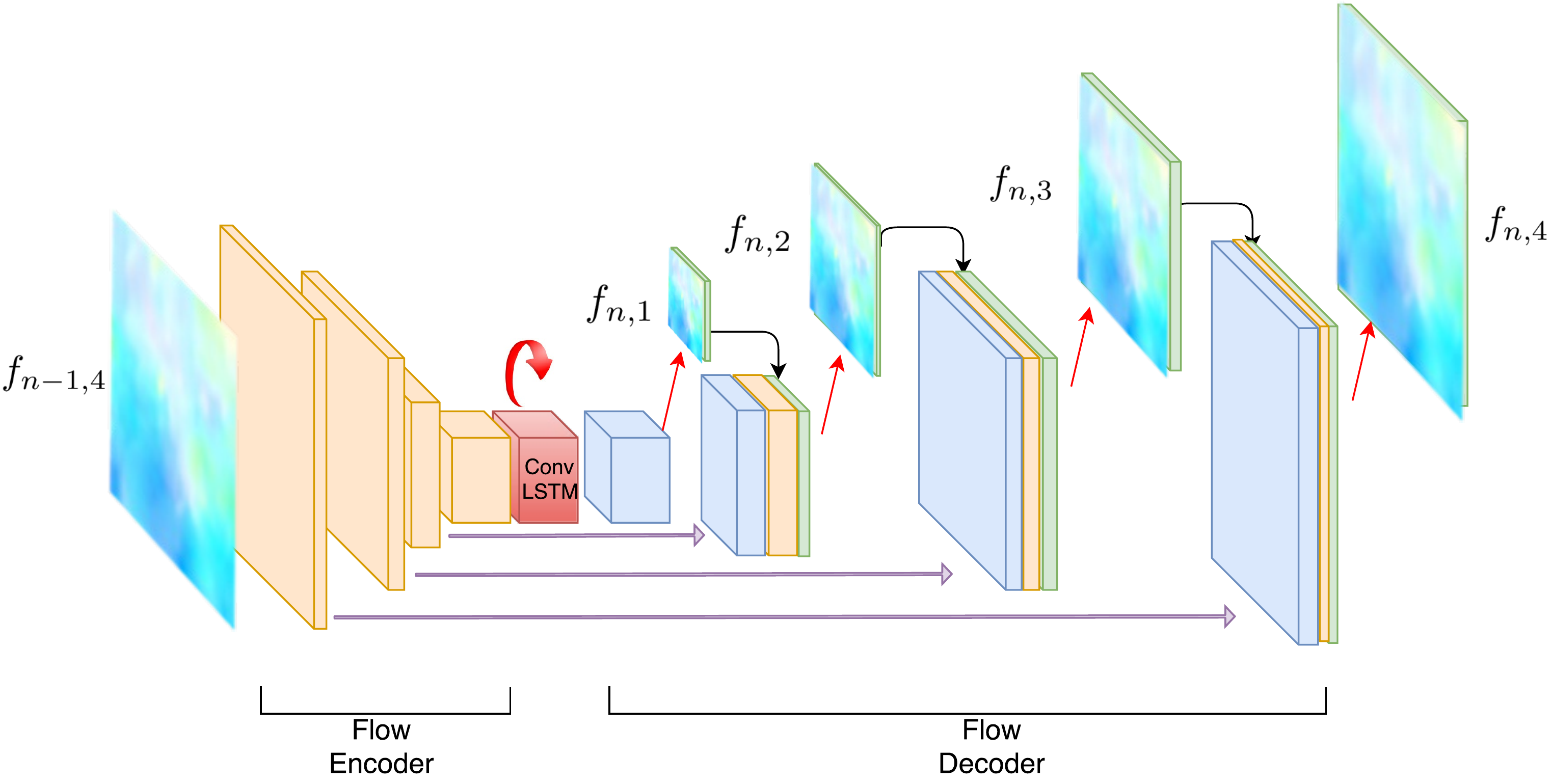}
\end{center}
\vspace{-4mm}
   \caption{Our Recurrent Video Decoder (RVD). This module recurrently generates optical flows which are warped to transform the sharp frame. Flows are estimated at 4 different scales.}
\label{fig:RVD}
\end{figure}

The task of RVD is to construct a sequence of frames using the motion representation provided by RVE and the (known) central frame ($x_{\lfloor\frac{N}{2}\rfloor}$) of the sequence. The RVD contains a flow encoder which utilizes a structure similar to the RVE. Instead of accepting images, it accepts optical flows. The flow encoding is fed to a ConvLSTM cell whose first hidden state is initialized with the last hidden state $h_{e,N}$ of the RVE. To estimate optical flows for a time-step, the output of the ConvLSTM cell is passed to a Flow decoder network ($F_D$). The flow estimated by $F_D$ at each time-step is fed to a transformer module ($T$) which returns the estimated frame $\hat{x}_{n}$. The descriptions of $F_D$ and $T$ are provided below.
\\
\textbf{Flow Decoder ($F_D$)}:
Realizing that the flow at current step is related to the previous one, 
we perform recurrence on optical flows for consecutive frames. The design of $F_D$ is illustrated in Fig. \ref{fig:RVD}. 
$F_D$ accepts the output of ConvLSTM unit at any time-step and generates a flow-map. For robust estimation, we further perform estimation of flow at multiple scales using deconvolution (deconv) layers which ``unpool'' the feature maps and increase the spatial dimensions by a factor of $2$. Inspired by \cite{ronneberger2015u}, we make use of skip connections between the layers of flow encoder and $F_D$.
All deconv operations use $ 4\times 4$ filters and the convolutional operations use $3\times3$ filters. The output of the ConvLSTM cell is passed through a convolutional layer to estimate the flow $f_{n,1}$. The cell output is also passed through a deconv layer before being concatenated with the upsampled $f_{n,1}$ and the corresponding feature-map coming from the encoder, to obtain a hybrid feature map at that scale. As shown in Fig. \ref{fig:RVD}, this process is repeated $3$ more times to obtain the flow maps at subsequently higher scales ($f_{n,2...4}$).
\\
\textbf{Transformer($T$)}: This generates a new frame by transforming a sharp frame using the output returned by $F_D$. It is a modified version of the Spatial Transformer Layer \cite{jaderberg2015spatial}, which comprises of a grid generator followed by a sampler. Instead of a single transformation for the entire image (as originally proposed in \cite{jaderberg2015spatial}), $T$ accepts one transformation per pixel. Since we focus on learning features for motion prediction, it provides immediate feedback on the flow map predicted by the optical flow generation layers. Effectively, the RVD function can be summarized as follows:
\vspace{-1mm}
\begin{eqnarray}
h^{dec}_1 = h^{enc}_N \\
h^{dec}_n,f_{n,1..4} = G(h^{dec}_{n-1},f_{n-1,4})\\
\hat{x}_{n,1..4} = T(x_{\lfloor\frac{N}{2}\rfloor},f_{n,1..4})
\end{eqnarray}

\noindent for $n$ $\in$ [1,N] where $h^{dec}_n$ is decoder hidden state, $f_{n,1..4}$ are flows predicted at $n$ and $\hat{x}_{n,1..4}$ are sharp frames predicted at different scales and $G$ refers to a recurrent cell of RVD.

\subsection{Blurred Image Encoder (BIE)}
\label{sec:bie}

We make use of the trained encoder-decoder couplet to solve the task of extracting video from a blurred image. 
We advocate a novel strategy of utilizing spatio-temporal embeddings to guide the training of a CNN. The trained decoder has learnt to generate optical flow for all time-steps from the encoder's hidden state. We make use of this proxy network to solve the task of blurred image to video generation.

The use of optical flow recurrence enables our network to prefer temporally consistent sequences, which preempts it from returning arbitrarily ordered frames. However, directional ambiguity stays.
For a scene with multiple objects, the ambiguity becomes more pronounced as each object can have its own independent motion. The BIE is connected with the pre-trained RVD and the pair is trained (RVD is fine-tuned) using a combination of ordering-invariant frame reconstruction loss and spatial motion smoothness loss over the RVD outputs (described later). No such ambiguity exists in the video autoencoder since the RVD has to exactly reproduce the video which is fed to RVE.

The BIE is implemented as a CNN which specializes in extracting motion features from a blurred image (we experimentally found that feeding the central sharp frame along with the blurred image improves its performance). The BIE is tasked to extract the sequential motion in the image by capturing local motion, e.g. at the smeared edges in the image. Moreover, the generated encoding should be such that the RVD can reconstruct motion trajectories. The BIE has 7 convolutional layers with kernel sizes as shown in Fig. \ref{fig:BIEnRVE}(b). Each layer (except the last) is followed by batch-normalization and leaky ReLU non-linearity.

\subsection{Cost Function}
\label{sec:Cost Function}
Both our network pairs (RVE-RVD and BIE-RVD) are trained by calculating the cost on the flows and frames estimated by the RVD. Since RVD implicitly estimates optical flows, we utilize a cost function motivated by learning-free variational method \cite{brox2011large} which resembles the original formulation of \cite{horn1981determining} to impose flow smoothness. At each time step, the data loss measures the discrepancy between intensities of target frame and the output of transformation layer (obtained using the the predicted optical flow field). The smoothness cost is in the form of total variation-loss on the estimated flow-maps: $TV(s)= \sum \vert{\nabla_x} s \vert + \vert{\nabla_y} s\vert $. 
\\
\begin{figure}
\begin{center}
   \includegraphics[width=0.5\textwidth]{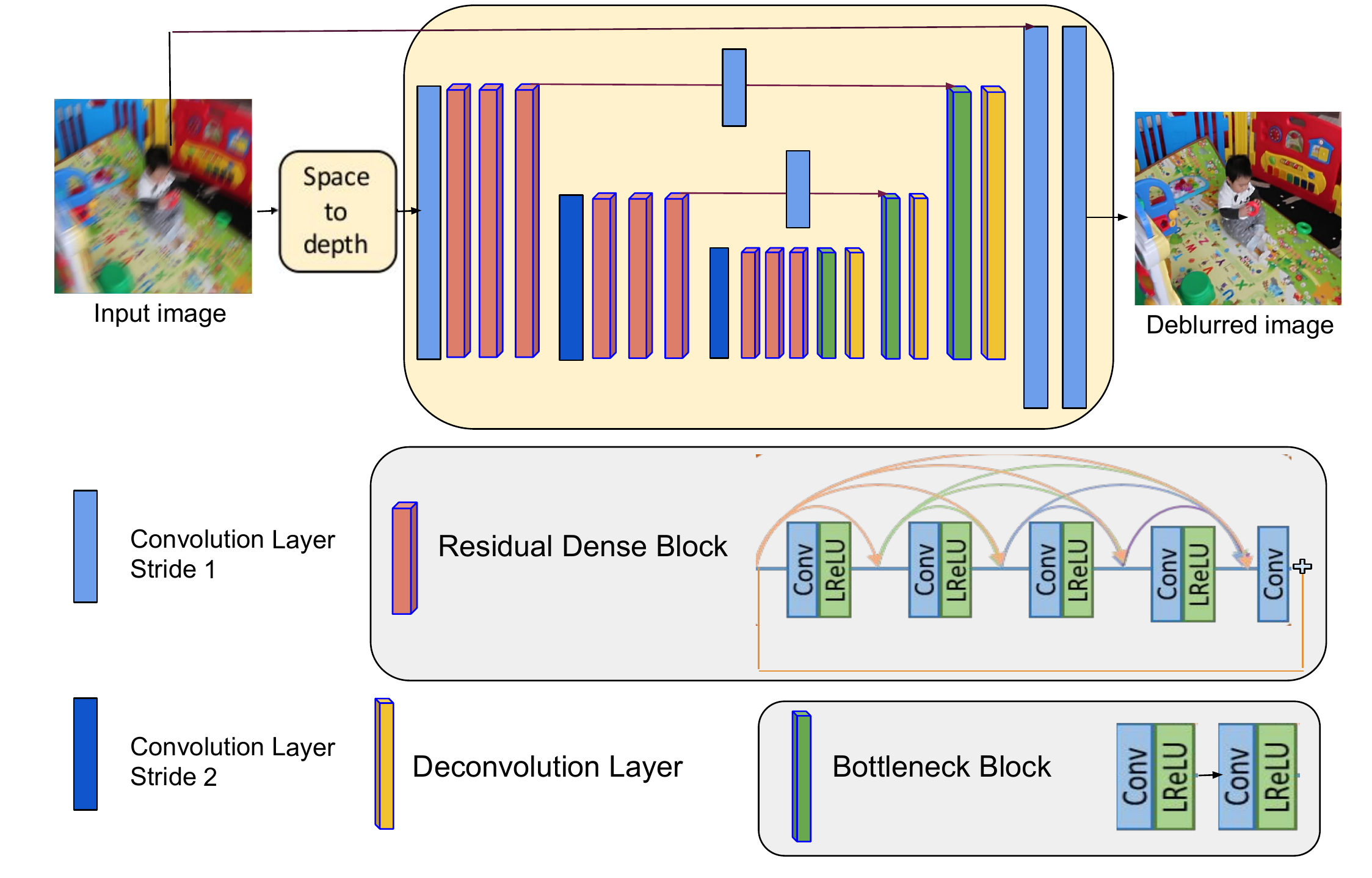}
\end{center}
\vspace{-4mm}
   \caption{An overview of our dense deblurring architecture which we utilize to estimate the central sharp frame. It follows an encoder-decoder design with residual-dense blocks, bottleneck blocks, and skip connections present at 3 different sub-scales.}
\label{fig:SID}
\end{figure}
\\
\textbf{Coarse-to-Fine}:  
Motivated by the approach employed in FlowNet \cite{dosovitskiy2015flownet}, we improve our network's accuracy by estimating the flow-maps and frames in a coarse-to-fine manner. At each time-step, four loss terms are calculated using four optical flows $f_{n,1..4}$ predicted at sizes which are $(\frac{1}{8},\frac{1}{4},\frac{1}{2},1)^{th}$ fraction of the original image resolution and applied on the corresponding down-sampled central frame using the transformation layers. 
Reconstruction losses are calculated at each scale using suitably down-sampled ground truth videos.  Effectively, we use a weighted sum of loss to guide the information flow over the network, which can be expressed as 

\begin{equation}
\mathcal{L} = \sum_{j=1}^{4} \lambda_j \left[  \mathcal{L}_j + \sum_{n=1}^{N} \mu TV(f_{n,j})) \right] 
\end{equation}
For RVE-RVD training, the data term we use is
\begin{equation}
\mathcal{L}_j = \sum_{n=1}^N \Big| \hat{x}_{n,j} - x_{n,j} \Big|_1
\end{equation}

As mentioned in section \ref{sec:bie}, training of BIE-RVD requires a loss term that preempts the network from penalizing a video which correctly explains the blurred image but does not match the available ground truth. Following \cite{jin2018learning}, we use loss function

\begin{align}
\mathcal{L}_j= \sum_{n=1}^\frac{N}{2} &\textstyle \Big| |\hat{x}_{n,j} + \hat{x}_{N-n,j}| - |x_{n,j} + x_{N-n,j}| \Big|_1 \notag \\ 
 +&\textstyle\Big| |\hat{x}_{n,j} - \hat{x}_{N-n,j}| - |x_{n,j} - x_{N-n,j}| \Big|_1
\label{eq:SumDiff_blurryInd}
\end{align} 

\noindent Here, $j$ represents the scale, $n$ represents time-step, $\mu$ is the regularization weight for total-variation loss empirically set to $0.02$. The relative weights $\lambda_j$s for each scale were adopted according to the loss weight suggested in \cite{mayer2016large}. 

\subsection{Deblurring Module (DM)}

We propose an independent network for deblurring the motion blurred observation. The estimated sharp frame is fed to both BIE and RVD during testing. 

  Recent works on image restoration have proposed end-to-end trainable networks which require labeled pairs of degraded and sharp images. Among them, \cite{nah2017deep,tao2018scale} have achieved promising results using multi-scale CNN composed of residual connections. We explore a more effective network architecture which is inspired by prior methods that use multi-level and multi-scale features. Our high-level design is similar to that of U-Net~\cite{ronneberger2015u}, which has been used extensively for preserving global context information in various image-to-image tasks~\cite{isola2017image}. Based on the observation that increase in number of layers and connections across them leads to a boost in feature extraction capability, the encoder structure of our network utilizes a cascade of Residual Dense Blocks (RDB)~\cite{zhang2018residual} instead of convolutional layers. An RDB is a cascade of convolutional layers connected through a rich set of residual and concatenation connections which immensely improves feature extraction capability by reusing features across multiple layers.  Inclusion of such connections maximizes information flow along the intermediate layers and results in better convergence. These units efficiently learn deeper and more complex features than a network with residual connections (which have been used extensively in recent deblurring methods\cite{nah2017deep,kupyn2017deblurgan,tao2018scale,jin2018learning}), while requiring fewer parameters.

Our proposed deblurring architecture is depicted in Fig. \ref{fig:SID}. The decoder part of our network contains $3$ pairs of up-sampling blocks to gradually enlarge the spatial resolution of feature maps. Each up-sampling block contains a bottleneck layer~\cite{jegou2017one} followed by a deconvolution layer. Each convolution layer (except the last) is followed by a non-linearity. Similar to U-Net, features corresponding to the same dimension in encoder and decoder are merged with the help of projection layers. The output of the final up-sampling block is passed through two additional convolutional layers to reconstruct the output sharp image. Our network uses an asymmetric encoder-decoder architecture, where the network capacity becomes higher benefiting from the dense connections. 

Further, we optimize the inference time of the network by performing computationally intensive operations on features at lower spatial resolution. This also reduces memory footprint while increasing the receptive field. Specifically, before feeding the input blurred image to encoder, we map the image to a lower resolution space using space-to-depth transformation. Following \cite{lim2017enhanced,nah2017deep}, we omit normalization layers for stable training, better generalization and reduced computational complexity and memory usage. To further improve performance, we also exploit residual scaling \cite{lim2017enhanced}.

\section{Experiments}

In this section, we carry out quantitative and qualitative comparisons of our approach with state-of-the-art methods for deblurring as well as video extraction tasks.  

\subsection{Implementation Details} 
We prepared our training data from GoPro dataset \cite{nah2017deep}, following standard train-test split, wherein $22$ full videos were used for creating training sets and $11$ full videos were reserved for validation and testing. 
Each blurred image is produced by averaging 9 successive latent frames. Such an averaging simulates a photo taken at approximately $26$ fps, while the corresponding sharp image shutter speed is $1/240$. We extract $256\times256$ patches from these image sequences for training. 
Finally, our dataset is composed of $10^5$ sets, each containing $N=9$ sharp frames and the corresponding blurred image $x_B$.
We perform data augmentation by random horizontal flipping and zooming by a factor in the range [$0.2,2$].
The network is trained using Adam optimizer with learning rate $1 \times 10^{-4}$. The batch size was set to 10 and the training of our video-autoencoder took $5 \times 10^4$ iterations to converge. We then train the BIE-RVD pair with the same training configuration and reduce the learning rate for RVD parameters to $2 \times 10^{-5}$, for stable training.

For training and evaluating our single image deblurring network, we utilized the same train-test split of the GoPro dataset \cite{nah2017deep} as recent deblurring methods \cite{nah2017deep}\cite{tao2018scale}. The batch size was set to 16 and the entire training took $4.5 \times 10^5$ iterations to converge.
\vspace{-3mm}
\subsection{Results for Single Image Deblurring }
\begin{figure*}%[b!]
	\scriptsize
	\centering
	\begin{tabular}{cc}	
		\begin{adjustbox}{valign=t}
		\tiny
			\begin{tabular}{c}
				\includegraphics[width=0.22\textwidth]{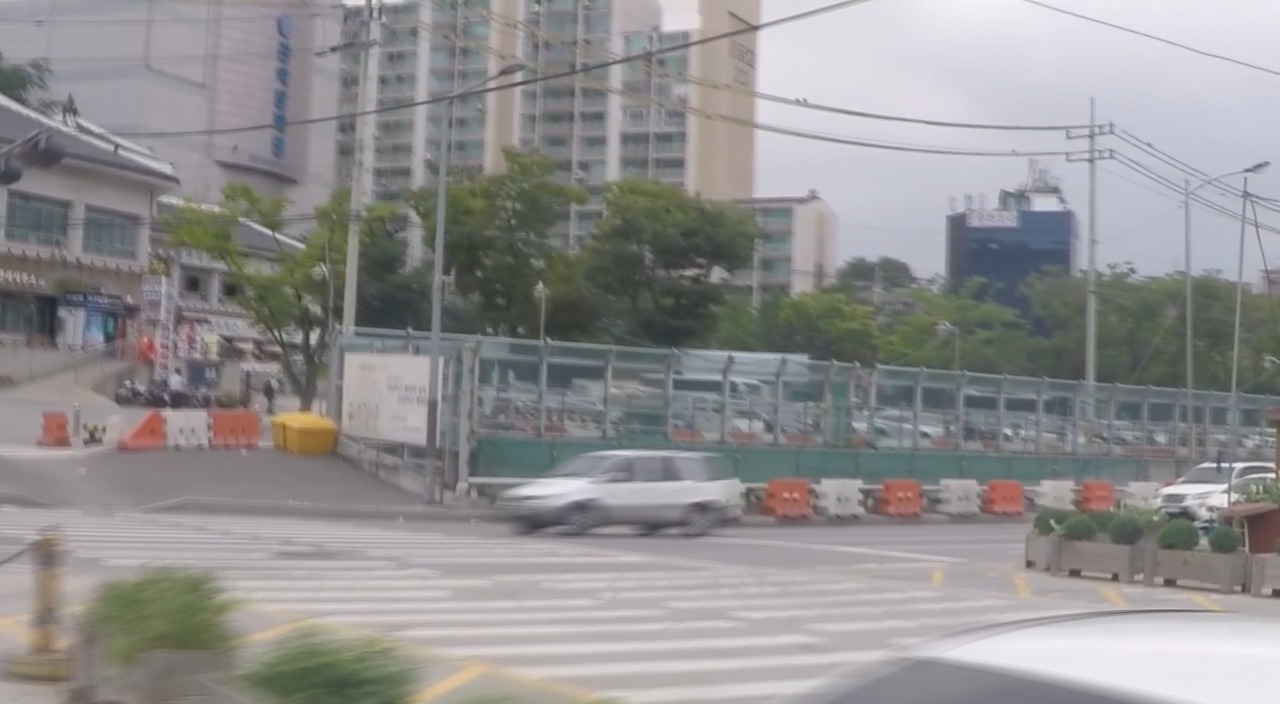}
				\\
			%	Blurred Image
				%\textsc{B100}: img_092
				
			\end{tabular}
		\end{adjustbox}
		\hspace{-1.3mm}
		\begin{adjustbox}{valign=t}
		\tiny
			\begin{tabular}{cccccc}
				\includegraphics[bb=490 170 740 280,clip=True,width=\widthscalefive \textwidth]{deblurring/000105_blurred.jpg} & 
				\includegraphics[bb=499 178 749 288,clip=True,width=\widthscalefive \textwidth]{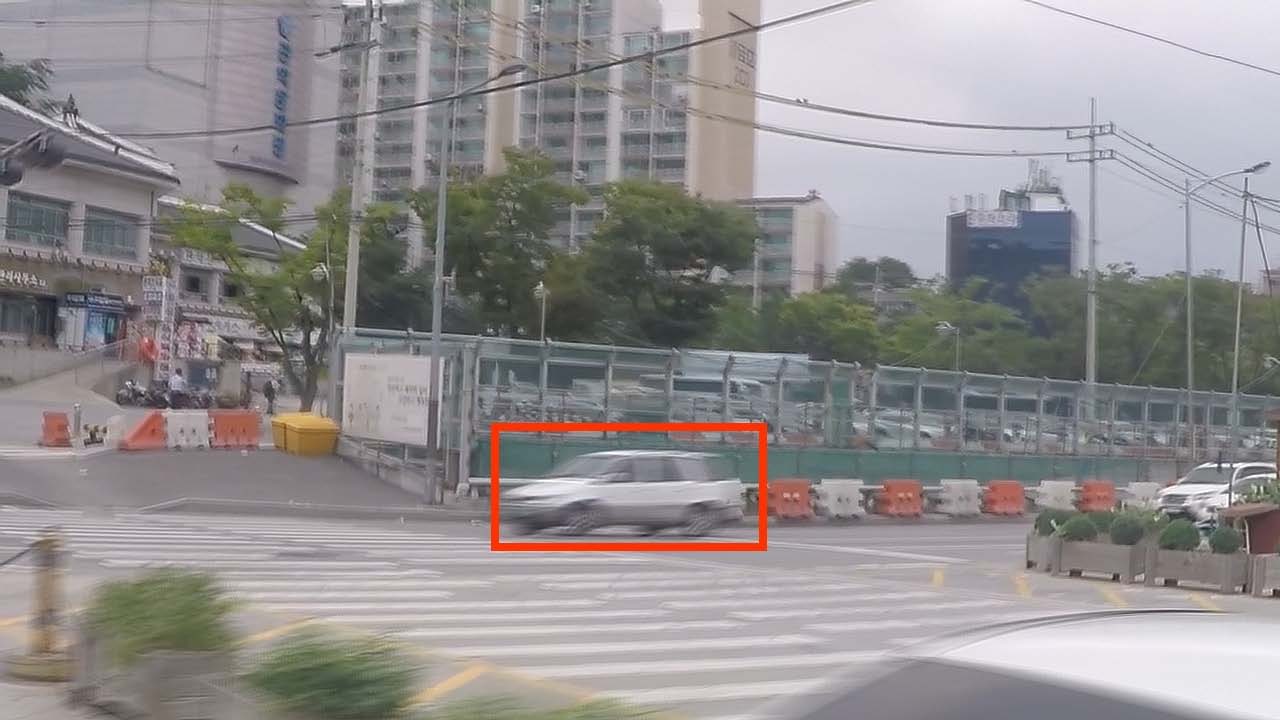} & %\hspace{\fsdttwofig} &
				\includegraphics[bb=490 180 740 290,clip=True,width=\widthscalefive \textwidth]{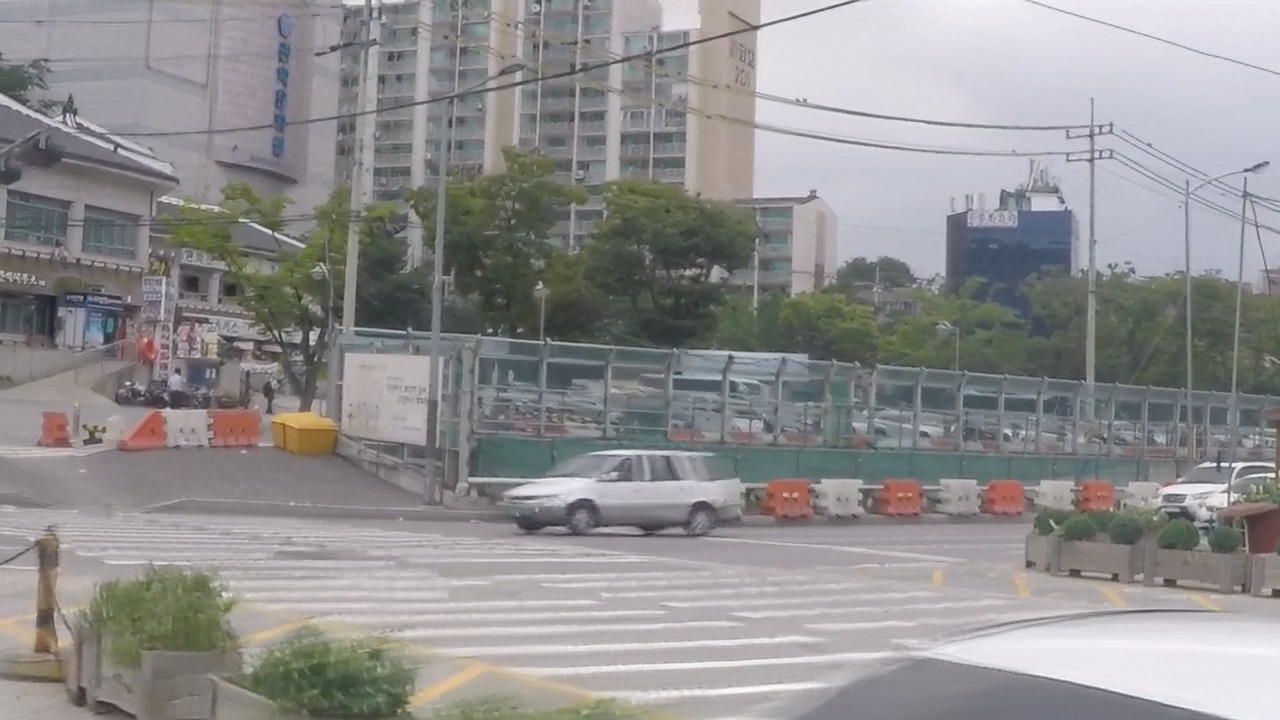} & %\hspace{\fsdttwofig} &
				\includegraphics[bb=490 170 740 280,clip=True,width=\widthscalefive \textwidth]{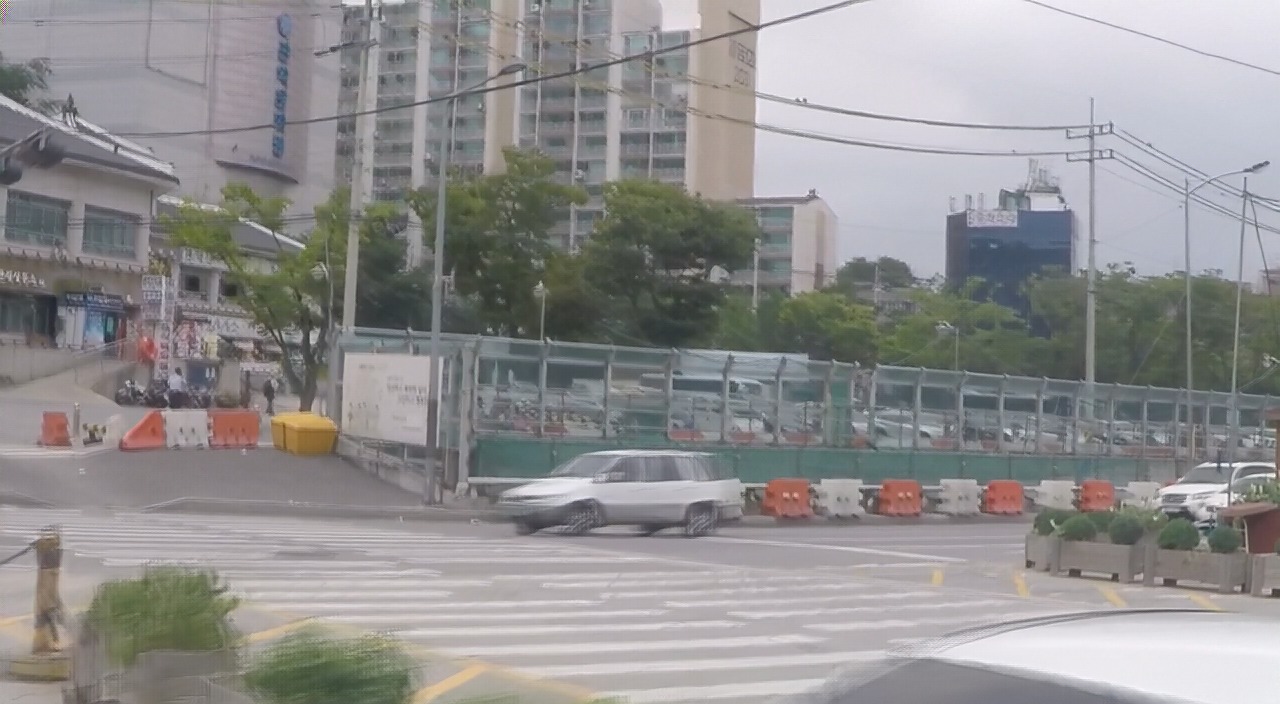} & %\hspace{\fsdttwofig} &
								\includegraphics[bb=490 170 740 280,clip=True,width=\widthscalefive \textwidth]{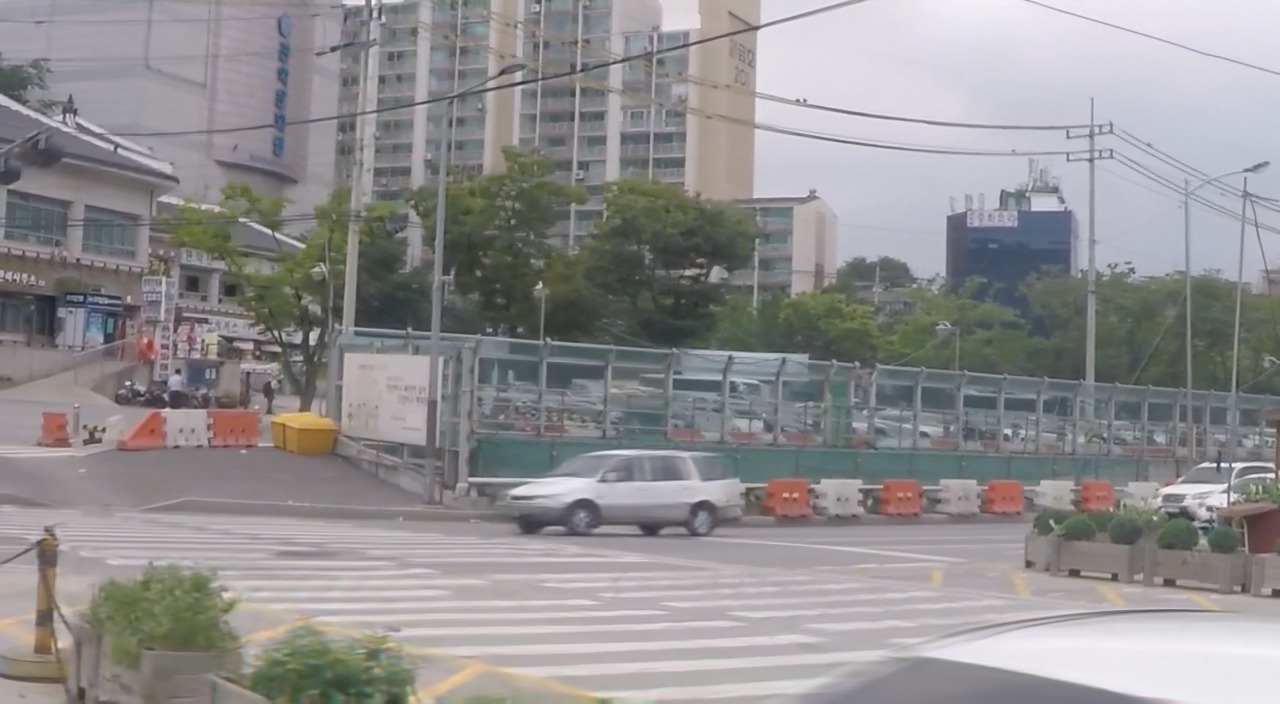} & %\hspace{\fsdttwofig} &
				\includegraphics[bb=490 170 740 280,clip=True,width=\widthscalefive \textwidth]{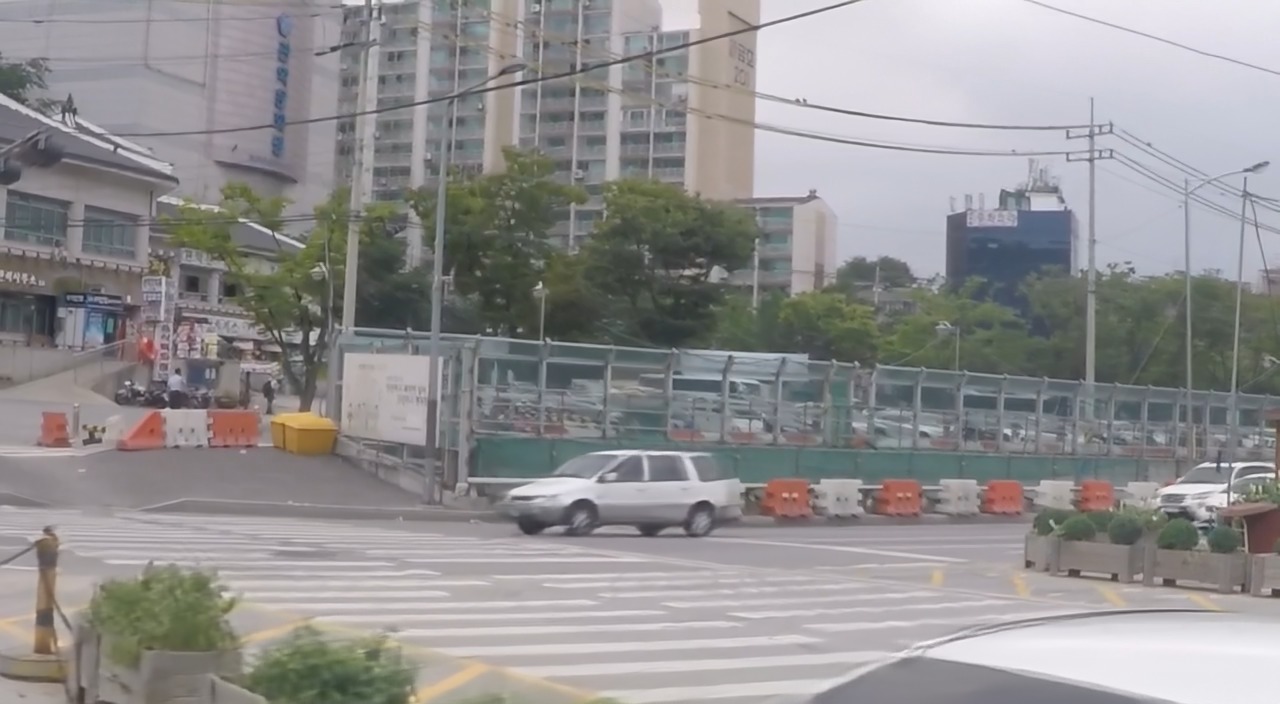}
				\\
				\includegraphics[bb=680 1 1040 150,clip=True,width=\widthscalefive \textwidth]{deblurring/000105_blurred.jpg} & %\hspace{\fsdttwofig} &
				\includegraphics[bb=680 1 1040 150,clip=True,width=\widthscalefive \textwidth]{deblurring/000105_whyte.jpg} & %\hspace{\fsdttwofig} &
				\includegraphics[bb=680 1 1040 150,clip=True,width=\widthscalefive \textwidth]{deblurring/000105_nah.jpg} & %\hspace{\fsdttwofig} &
				\includegraphics[bb=680 1 1040 150,clip=True,width=\widthscalefive \textwidth]{deblurring/000105_deblurgan.jpg} & %\hspace{\fsdttwofig} &
								\includegraphics[bb=680 1 1040 150 280,clip=True,width=\widthscalefive \textwidth]{deblurring/000105_srn.jpg} & %\hspace{\fsdttwofig} &
				\includegraphics[bb=680 1 1040 150,clip=True,width=\widthscalefive \textwidth]{deblurring/000105_ours.jpg} 
				\\
			\end{tabular}
		\end{adjustbox}
				\\			
		\begin{adjustbox}{valign=t}
		\tiny
			\begin{tabular}{c}
				\includegraphics[width=0.22\textwidth]{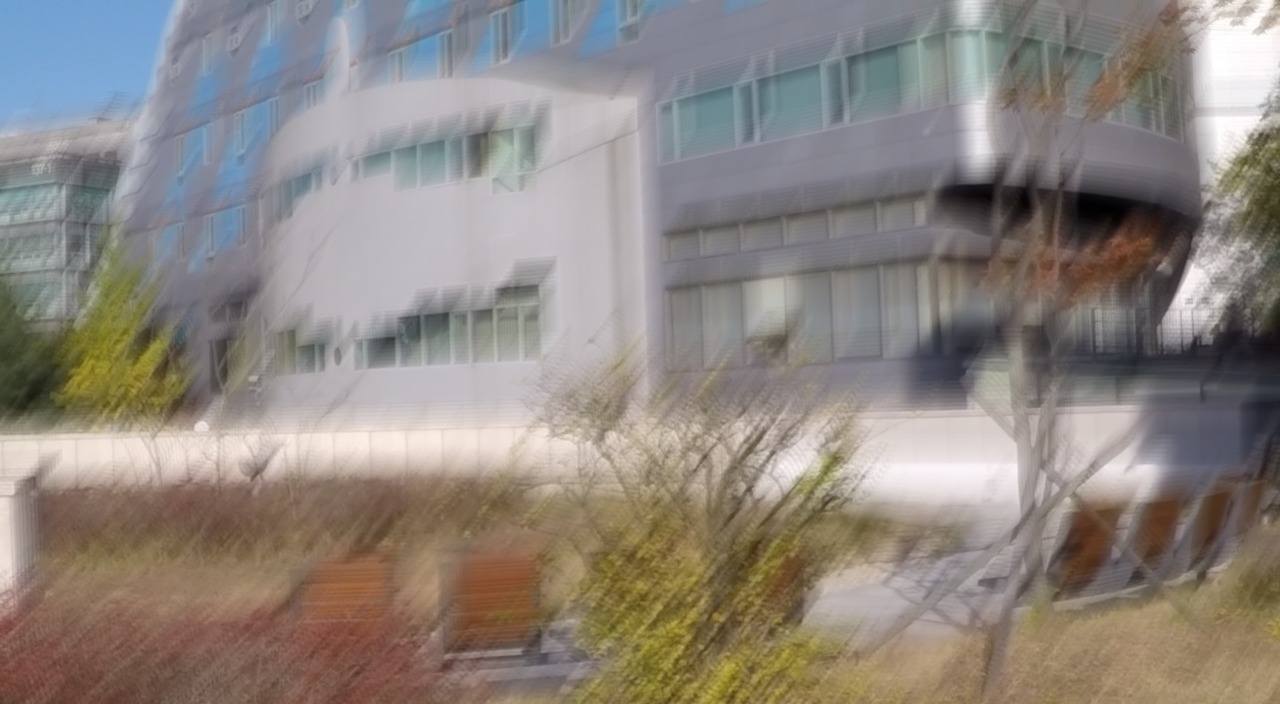}
				\\
				Blurred Image
			\end{tabular}
		\end{adjustbox}
		\hspace{-1.3mm}
		\begin{adjustbox}{valign=t}
		\tiny
			\begin{tabular}{cccccc}
				\includegraphics[bb=480 1 880 200,clip=True,width=\widthscalefive \textwidth]{deblurring/000045_blurred.jpg} & %\hspace{\fsdttwofig} &
				\includegraphics[bb=480 1 880 200,clip=True,width=\widthscalefive \textwidth]{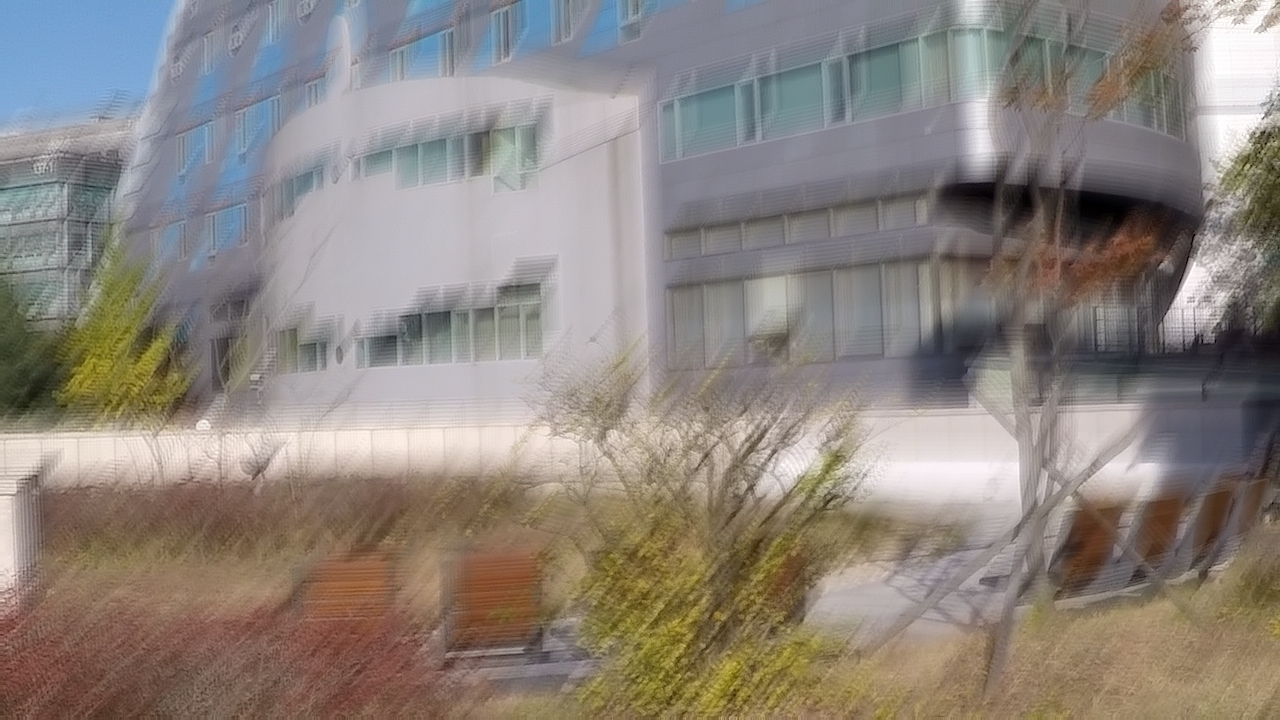} & %\hspace{\fsdttwofig} &
				\includegraphics[bb=480 1 880 200,clip=True,width=\widthscalefive \textwidth]{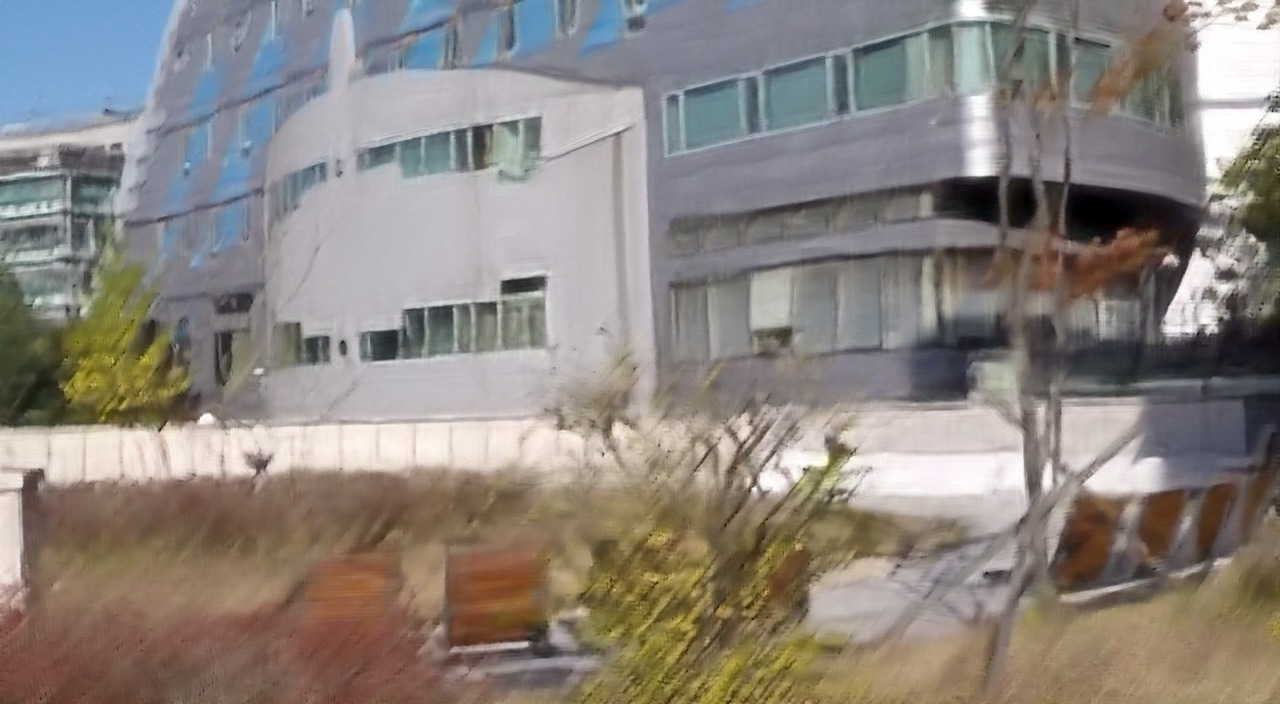} & %\hspace{\fsdttwofig} &
				\includegraphics[bb=480 1 880 200,clip=True,width=\widthscalefive \textwidth]{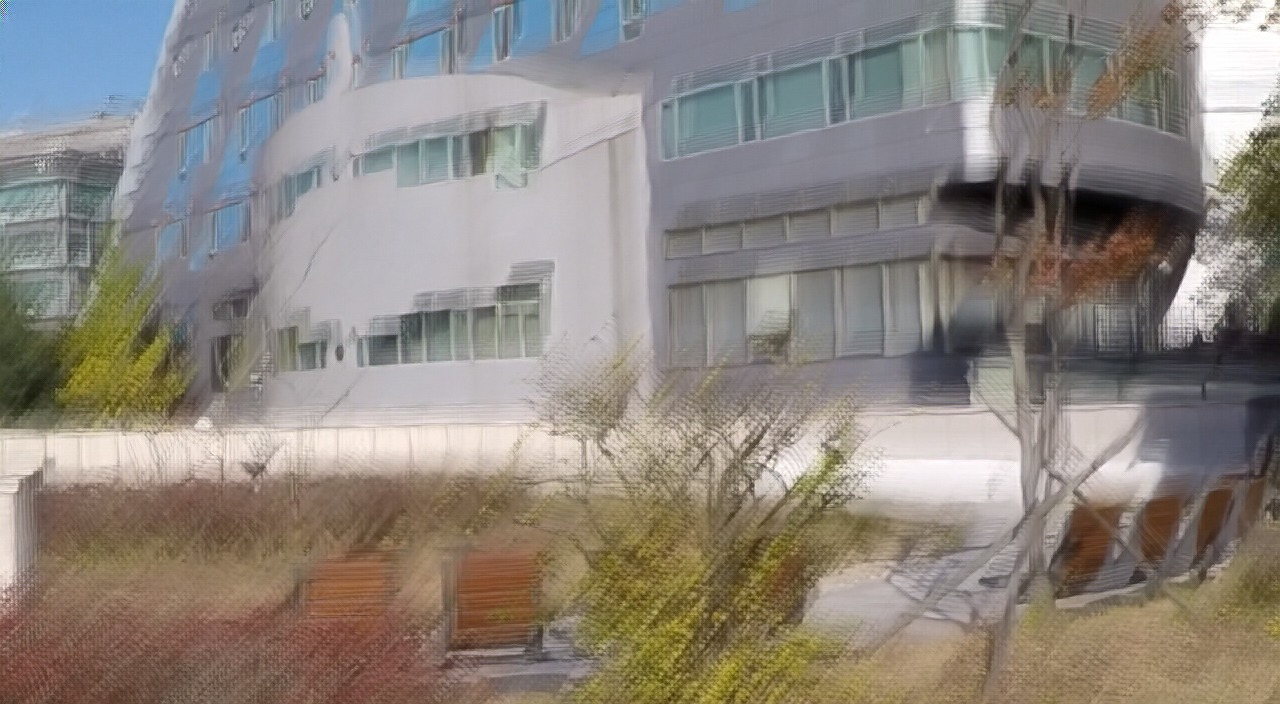} & %\hspace{\fsdttwofig} &
								\includegraphics[bb=480 1 880 200 280,clip=True,width=\widthscalefive \textwidth]{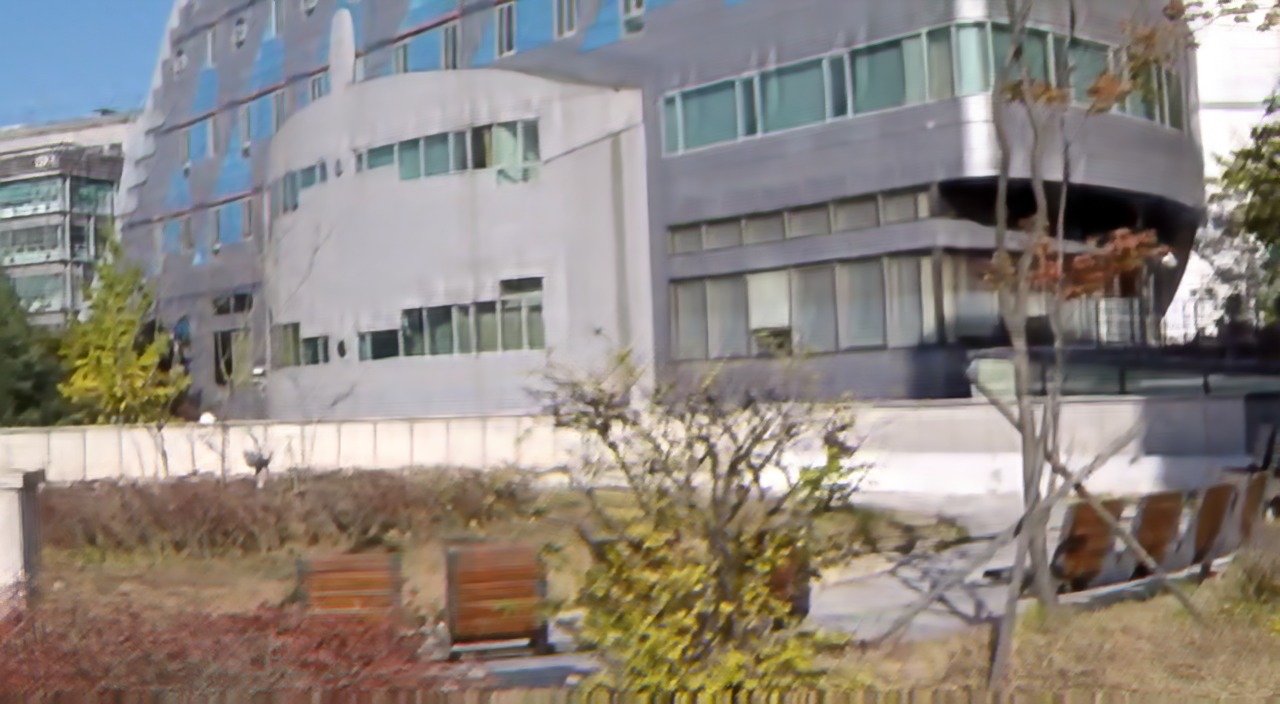} & %\hspace{\fsdttwofig} &3
				\includegraphics[bb=480 1 880 200,clip=True,width=\widthscalefive \textwidth]{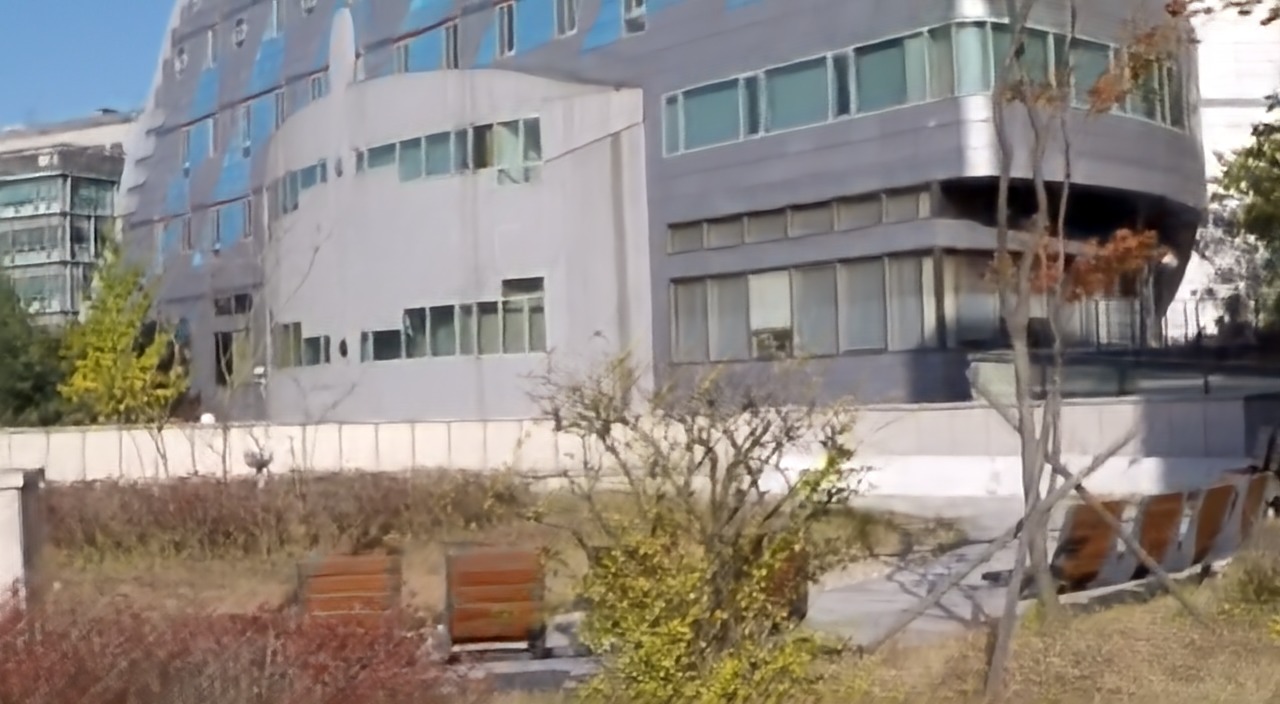} 				\vspace{-3mm}

				\\ 
				\includegraphics[bb=380 400 980 780,clip=True,width=\widthscalefive \textwidth]{deblurring/000045_blurred.jpg} & %\hspace{\fsdttwofig} &
				\includegraphics[bb=380 400 980 780,clip=True,width=\widthscalefive \textwidth]{deblurring/000045_whyte.jpg} & %\hspace{\fsdttwofig} &
				\includegraphics[bb=380 400 980 780,clip=True,width=\widthscalefive \textwidth]{deblurring/000045_nah.jpg} & %\hspace{\fsdttwofig} &
				\includegraphics[bb=380 400 980 780,clip=True,width=\widthscalefive \textwidth]{deblurring/000045_deblurgan.jpg} & %\hspace{\fsdttwofig} &
								\includegraphics[bb=380 400 980 780 280,clip=True,width=\widthscalefive \textwidth]{deblurring/000045_srn.jpg} & %\hspace{\fsdttwofig} &
				\includegraphics[bb=380 400 980 780,clip=True,width=\widthscalefive \textwidth]{deblurring/000045_ours.jpg} 
				\\ 
				Blurred patch& %\hspace{\fsdttwofig} &
				Whyte \etal~\cite{whyte2012non} & %\hspace{\fsdttwofig} &
				Nah \etal~\cite{nah2017deep} & %\hspace{\fsdttwofig} &
				DelurGAN~\cite{kupyn2017deblurgan}& % \hspace{\fsdttwofig} &
				SRN~\cite{tao2018scale}& %\hspace{\fsdttwofig} &
				Ours%\hspace{\fsdttwofig} &
				\\
			\end{tabular}
		\end{adjustbox}	
	\end{tabular}
	\caption{Visual comparisons of deblurring results on test dataset~\cite{nah2017deep} (best view in high resolutions).}
\label{fig:comp_deep}
\end{figure*}

We evaluated the efficacy of our network (DM shown in Fig. \ref{fig:SID}) for the intermediate task of deblurring, both quantitatively and qualitatively on $1100$ test images (resolution $1280\times704$) from the GoPro dataset \cite{nah2017deep}. The method of \cite{whyte2012non} is selected as representative traditional method for non-uniform blur. We also compare our performance with deep networks \cite{nah2017deep,kupyn2017deblurgan,tao2018scale}. All the codes were downloaded from the respective authors' websites. Quantitative and qualitative comparisons are presented in Table \ref{TableGopro} and Fig. \ref{fig:comp_deep}, respectively. Since traditional method of \cite{whyte2012non} cannot model combined effects of general camera shake and object motion, it fails to faithfully restore most of the images in the test-set. On the other hand, the method of \cite{kupyn2017deblurgan} trains a residual network containing instance-normalization layers using a mixture of deep-feature losses and adversarial losses, but leads to suboptimal performance on images containing large blur. The methods \cite{nah2017deep,tao2018scale} use a multi-scale strategy to improve capability to handle large blur, but fail in challenging situations. Fig. \ref{fig:comp_deep} shows that results of prior works suffer from incomplete deblurring or ringing artifacts. In contrast, our network is able to restore scene details more faithfully, while being $20$ times faster than the nearest competitor~\cite{tao2018scale}. These improvements are also reflected in the quantitative values presented in the table.

\setlength{\tabcolsep}{1.4pt}
\begin{table}[]
\centering
\begin{tabular}{|c|c|c|c|c|c|c|c|c|}
\hline

Method & \cite{xu2013unnatural} & \cite{whyte2012non} & \cite{sun2015learning} & \cite{gong2017motion} & \cite{nah2017deep} & \cite{kupyn2017deblurgan} & \cite{tao2018scale} & Ours \\
\hline
\small{PSNR(dB)} & 21 & 24.6 & 24.5 &  26.4  & 28.9 & 27.2 & 30.10 & \textbf{30.58} \\
\small{SSIM} & 0.740 & 0.845 & 0.851 & 0.863 & 0.911 & 0.905 & 0.933 & \textbf{0.941}\\
\small{Time (s)} & 3800 & 700 & 1500 & 1200 & 6 & 0.8 & 0.4 & \textbf{0.02} \\
\small{Size(MB)} & - & - & 54.1 & 41.2 & 300 & 45.6 & 27.5 & \textbf{17.9} \\
\hline
\end{tabular}
\vspace{0mm}
\caption{Performance comparison of our deblurring network with existing methods on the benchmark dataset \cite{nah2017deep}.
\vspace{-4mm}
\label{TableGopro}}
\end{table}
\setlength{\tabcolsep}{1.4pt}

 \begin{figure*}[t]
 \begin{center}
 \begin{tabular}{cccccc}
   \includegraphics[width=\widthscalefour\textwidth]{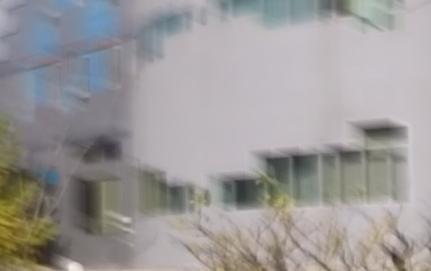} &
   \includegraphics[width=\widthscalefour\textwidth]{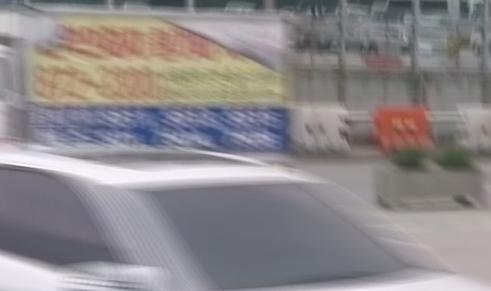} &
      \includegraphics[width=\widthscalefour\textwidth]{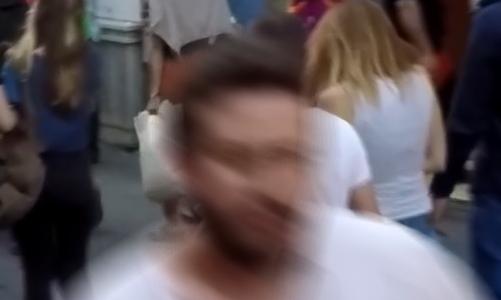} &
\includegraphics[width=\widthscalefour\textwidth]{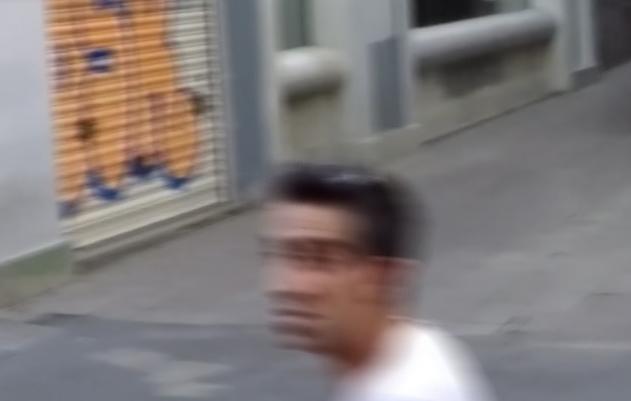}  &
   \includegraphics[width=\widthscalefour\textwidth]{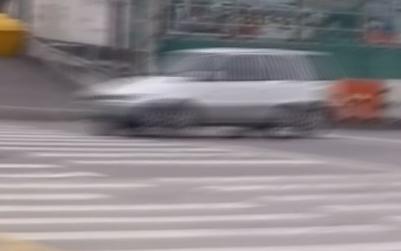} &       
          \includegraphics[width=\widthscalefour\textwidth]{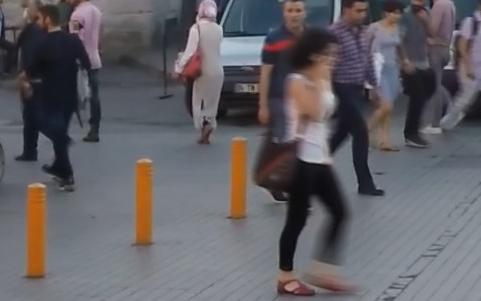}
           \\

   \animategraphics[width=\widthscalefour\textwidth   ,loop]{9}{"NEWcropped/ours1/"}{0}{8}&

   \animategraphics[width=\widthscalefour\textwidth    ,loop]{9}{"NEWcropped/ours3/"}{0}{8} &
      \animategraphics[width=\widthscalefour\textwidth    ,loop]{9}{"NEWcropped/ours4/"}{0}{8} &  
        \animategraphics[width=\widthscalefour\textwidth    ,loop]{9}{"NEWcropped/ours5/"}{0}{8}   &
   \animategraphics[width=\widthscalefour\textwidth    ,loop]{9}{"NEWcropped/ours2/"}{0}{8} &        
           \animategraphics[width=\widthscalefour\textwidth    ,loop]{9}{"NEWcropped/ours6/"}{0}{8}  \\

   \animategraphics[width=\widthscalefour\textwidth   ,loop]{8}{"NEWcropped/favaro1/000066-esti"}{1}{7}&
   \animategraphics[width=\widthscalefour\textwidth    ,loop]{8}{"NEWcropped/favaro3/000229-esti"}{1}{7} & 
      \animategraphics[width=\widthscalefour\textwidth    ,loop]{8}{"NEWcropped/favaro4/004039-esti"}{1}{7} & 
         \animategraphics[width=\widthscalefour\textwidth    ,loop]{8}{"NEWcropped/favaro5/003069-esti"}{1}{7} &  
   \animategraphics[width=\widthscalefour\textwidth    ,loop]{8}{"NEWcropped/favaro2/000108-esti"}{1}{7} &         
            \animategraphics[width=\widthscalefour\textwidth    ,loop]{8}{"NEWcropped/favaro6/000040-esti"}{1}{7}   \\         
   (a) & (b) & (c) & (d) & (e) & (f)
\end{tabular}
   \caption{Comparisons of our video extraction results with \cite{jin2018learning} on motion blurred images obtained from the test dataset of \cite{nah2017deep}. The first row shows the blurred images while the second and third rows show videos generated by our method and \cite{jin2018learning}, respectively. \emph{Videos can be viewed by clicking
on the image, when document is opened in Adobe Reader}.}  
\label{fig:gopropp}
\vspace{-5mm}
   \end{center}
\end{figure*}

\begin{figure*}[]
\begin{center}
\begin{tabular}{cccccc}
   \includegraphics[width=0.14\textwidth]{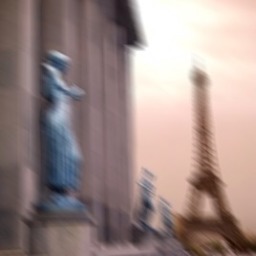} &
      \includegraphics[width=0.14\textwidth]{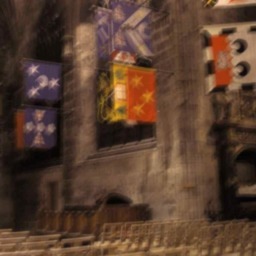} &
         \includegraphics[width=0.14\textwidth]{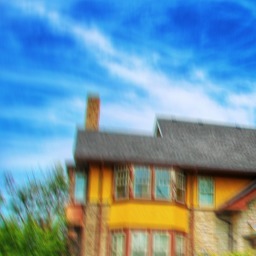} &
   \includegraphics[width=0.14\textwidth]{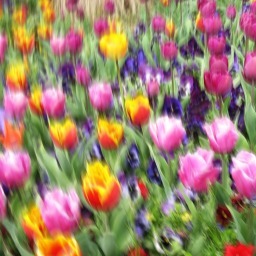} &
      \includegraphics[width=0.14\textwidth]{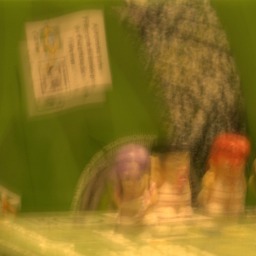}&
         \includegraphics[width=0.14\textwidth]{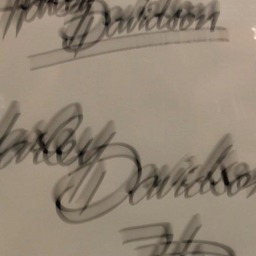}\\% &\linebreak%\vspace{1mm}
         \animategraphics[width=0.14\textwidth,loop]{8}{"3d/video/"}{2}{8}&
         \animategraphics[width=0.14\textwidth,loop]{8}{"kohler/video/"}{2}{8} &
         \animategraphics[width=0.14\textwidth,loop]{8}{"KaiSynth/house/video/"}{2}{8} &
            \animategraphics[width=0.14\textwidth,loop]{8}{"KaiSynth/flower/video/"}{2}{8} &
            \animategraphics[width=0.14\textwidth    ,loop]{8}{"KaiDataset/1/video/"}{2}{8}    &    
      \animategraphics[width=0.14\textwidth,loop]{8}{"KaiDataset/3/video/"}{2}{8}\\
   (a) & (b) & (c) & (d) & (e) & (f)
\end{tabular}  
\vspace{-1mm}
   \caption{Video generation from images blurred with global camera motion from datasets of \cite{gong2017motion},\cite{kohler2012recording} and \cite{lai2016comparative}. First row shows the blurred images. The generated videos using our method are shown in second row.}
\label{fig:cameramotion}
\end{center}
\vspace{-1.5em}
\end{figure*}

 \begin{figure*}[t]
 \begin{center}
 \begin{tabular}{ccccccc}
   \includegraphics[width=0.13\textwidth]{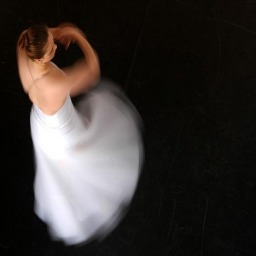} &
   \includegraphics[width=0.13\textwidth]{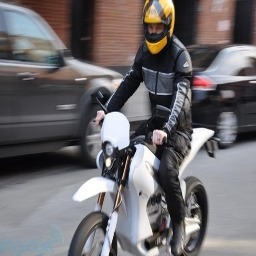} 8&

   \includegraphics[width=0.13\textwidth]{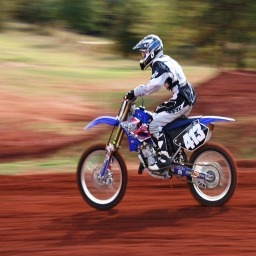} &

   \includegraphics[width=0.13\textwidth]{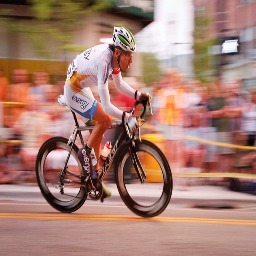} &
   \includegraphics[width=0.13\textwidth]{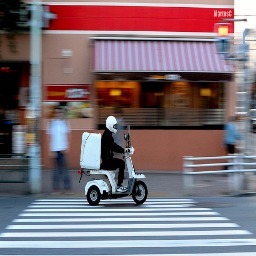} &   
      \includegraphics[width=0.13\textwidth]{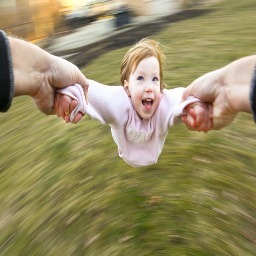} &
   \includegraphics[width=0.13\textwidth]{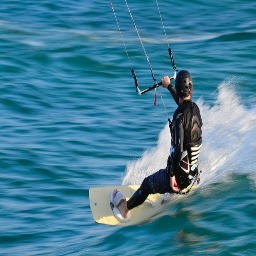}
           \\
\vspace{0mm}
   \animategraphics[width=0.13\textwidth    ,loop]{8}{"BlurDetectionDataset/im8/video/"}{2}{8}&
   \animategraphics[width=0.13\textwidth    ,loop]{8}{"BlurDetectionDataset/im12/video/"}{2}{8} &
   \animategraphics[width=0.13\textwidth    ,loop]{8}{"BlurDetectionDataset/im45/video/"}{2}{8} &
   \animategraphics[width=0.13\textwidth    ,loop]{8}{"BlurDetectionDataset/im256/video/"}{2}{8}&
   \animategraphics[width=0.13\textwidth   ,loop]{8}{"BlurDetectionDataset/im232/video/"}{2}{8} &
      \animategraphics[width=0.13\textwidth   ,loop]{8}{"BlurDetectionDataset/im272/video/"}{2}{8} &
   \animategraphics[width=0.13\textwidth   ,loop]{8}{"BlurDetectionDataset/im103/video/"}{2}{8}\\

\end{tabular}
   \caption{Video generation results on real motion blurred images from dataset of \cite{shi2014discriminative}. The first row shows the blurred images. Second row contains the extracted videos with our method.}
   \label{fig:blurdetection}
\vspace{-6mm}
   \end{center}

\end{figure*}

\subsection{Results and Comparisons for Video Extraction}
\label{favarocomparisons}

In Fig  \ref{fig:gopropp}, we give results on standard test blurred images from the dataset of \cite{nah2017deep}. Note that some of them suffer from significant blur. Fig. \ref{fig:gopropp}(a) shows an image of a planar scene which is blurred due to dominant camera motion. Fig. \ref{fig:gopropp}(b) shows a 3D scene blurred due to camera motion. Figs. \ref{fig:gopropp}(c-f) show results on blurred images with dynamic object motion. Observe that the videos generated by our approach are realistic and qualitatively consistent with the blur and depth of the scene, even when the foreground incurs large motion. Our network is able to reconstruct videos from blurred images with diverse motion and scene content.

In comparison, the results of \cite{jin2018learning} suffer from local errors in deblurring, inconsistent motion estimation, as well as color distortions. We have observed that in general the method of \cite{jin2018learning} fails in cases involving high blur as direct image regression becomes difficult for large motion. In contrast, we divide the overall problem into two sub-tasks of deblurring and motion extraction. This simplifies learning and yields improvement in deblurring quality as well as motion estimation. The color issue in \cite{jin2018learning} can be attributed to the design of their networks, wherein feature extraction and reconstruction branches are different for different color channels. Our method applies the same motion to each color channel. By having a single recurrent network to generate the video, our network can be directly trained to extract even higher number of frames ($>9$) without any design change or additional parameters. In contrast, \cite{jin2018learning} requires training of an additional network for each new pair of frames. Our overall architecture is more compact ($34$ MB vs $70$ MB) and much faster ($0.02$s vs $0.45$s for deblurring and $0.39$s vs $1.10$s for video generation) as compared to \cite{jin2018learning}. 

To perform quantitative comparisons with \cite{jin2018learning}, we also trained another version of our network on the restricted case of blurred images produced by averaging $7$ successive sharp frames. For testing, $250$ blurred images of resolution $1280 \times 704$ were created using the $11$ test videos from the dataset of \cite{nah2017deep}. We compared the videos estimated by the two methods using the ambiguity invariant loss function defined in Eq. \ref{eq:SumDiff_blurryInd}. The average error was found to be $49.06$ for \cite{jin2018learning} and $44.12$ for our method. Thus, even for the restricted case of small blur, our method performs favorably. Repeating the same experiment for $9$ frames (i.e. for large blur from the same test videos) led to an error of $48.24$ for our method, which is still less than the 7-frame error of \cite{jin2018learning}. We could not compute the 9-frame error for \cite{jin2018learning} as their network is rigidly designed for $7$ frames only.

\subsection{Additional Results on Video Extraction}

\noindent \textbf{Results on Camera Motion Dataset:}
For evaluating qualitative performance on videos with camera motion alone, we tested our network's ability to reconstruct videos from blurred images taken from datasets of \cite{gong2017motion}, \cite{kohler2012recording} and \cite{lai2016comparative}, which are commonly used for benchmarking deblurring techniques. Fig. \ref{fig:cameramotion}(a) shows the video obtained on a synthetically blurred image provided in \cite{gong2017motion}. Fig. \ref{fig:cameramotion}(b) shows result on an image from the dataset of \cite{kohler2012recording}. We can observe that the motion in the generated video conforms with the blur. The dataset \cite{lai2016comparative} consists of both synthetic and real images collected from various conventional prior works on deblurring. Figs. \ref{fig:cameramotion}(c-d) show our network's results on synthetically blurred images from this dataset using non-uniform camera motion. The examples in Figs. \ref{fig:cameramotion}(e-f) are real blurred images obtained from the same dataset. Our method is able to re-enact underlying motion quite well. 
\\
\textbf{Results on Blur Detection Dataset:}
In Fig. \ref{fig:blurdetection}, we show videos generated from real blurred images taken from the dataset of \cite{shi2014discriminative} which contains dynamic scenes. The results reaffirm that our network can sense direction and magnitude even in severely blurred images. 

\subsection{More Results and Ablation Studies}

Additional results and experiments to highlight the motivation for our design choices are given in the supplementary material. Specifically, for the video autoencoder, we study the effects of motion flow estimation (instead of direct intensity estimation) and the recurrent design. This is followed by an analysis on the influence of different loss functions. Regarding training of BIE, we study the effect of input sharp frame on its performance and also compare our two-stage strategy (BIE trained using pre-trained RVD) with the case where BIE and RVD are trained directly from scratch. We also include an analysis on variations in growth-rate and residual-dense connection topology on the training and test performance of our deblurring network. 

\section{Conclusions}
We introduced a new methodology for video generation from a single blurred image. We proposed a spatio-temporal video auto-encoder based on an end-to-end differentiable architecture that learns motion representation from sharp videos in a self-supervised manner. The network predicts a sequence of optical flows and employs them to transform a sharp central frame and return a smooth video. Using the trained video decoder, we trained a blurred image encoder to extract a representation from a single blurred image, that mimics the representation returned by the video encoder. This when fed to the decoder returns a plausible sharp video representing the action within the blurred image. We also proposed an efficient deblurring architecture composed of densely connected layers that yields state-of-the-art results. The potential of our work can be extended in a variety of directions including blur-based segmentation, video deblurring, video interpolation, action recognition etc.

{\small
\bibliographystyle{ieee}
\bibliography{egbib}
}

\end{document}